\pdfoutput=1

\documentclass[11pt]{article}

\usepackage[]{acl}

\usepackage{times}
\usepackage{latexsym}

\usepackage[T1]{fontenc}

\usepackage[utf8]{inputenc}

\usepackage{microtype}
\usepackage{multirow}
\usepackage{graphicx}
\usepackage{booktabs}
\title{Where to Go for the Holidays: Towards Mixed-Type Dialogs for Clarification of User Goals}

\author{
	Zeming Liu\textsuperscript{\rm 1}\thanks{\textit{ }\textit{ }Equal contribution.}\textit{ }\textit{ }\thanks{\textit{ }\textit{ }Mainly responsible for dataset collection during his internship at Baidu.},
	Jun Xu\textsuperscript{\rm 2}\footnotemark[1]\textit{ }\textit{ }\thanks{\textit{ }\textit{ }Corresponding author: Jun Xu.},
	Zeyang Lei\textsuperscript{\rm 2},
	Haifeng Wang\textsuperscript{\rm 2},
	Zheng-Yu Niu\textsuperscript{\rm 2},
	Hua Wu\textsuperscript{\rm 2}\\
	\textsuperscript{\rm 1}Research Center for Social Computing and Information Retrieval, \\Harbin Institute of Technology, Harbin, China\\
	\textsuperscript{\rm 2}Baidu Inc., Beijing, China\\ 
	zmliu@ir.hit.edu.cn, \{xujun03, leizeyang, wanghaifeng, niuzhengyu, wu\_hua\}@baidu.com 
}

\begin{document}
	\maketitle
	\begin{abstract}
		Most dialog systems posit that users have figured out clear and specific goals before starting an interaction. For example, users have determined the departure, the destination, and the travel time for booking a flight.  However, in many scenarios, limited by experience and knowledge, users may know what they need, but still struggle to figure out clear and specific goals by determining all the necessary slots. 
		
	    In this paper, we identify this challenge, and make a step forward by collecting a new human-to-human mixed-type dialog corpus. It contains 5k dialog sessions and 168k utterances for 4 dialog types and 5 domains. Within each session, an agent first provides user-goal-related knowledge to help figure out clear and specific goals, and then help achieve them. 
		
		Furthermore, we propose a mixed-type dialog model with a novel Prompt-based continual learning mechanism. Specifically, the mechanism enables the model to continually strengthen its ability on any specific type by utilizing existing dialog corpora effectively.
	\end{abstract}
	
\section{Introduction}
	One of the overarching goals of Artificial Intelligence is to build an intelligent agent that can generate coherent multi-turn dialogs to meet user needs/goals. Recently, multiple dialog agents have been launched, such as Echo and Siri. These agents usually position themselves as some kind of ``do engines'' that act under users' clear instructions. Specifically, they posit users have figured out clear and specific goals by determining all the necessary aspects or slots of their goals. For example, before booking a flight, a user has determined the departure, the destination and the travel time.

	However, such assumption can not hold in many real-world scenarios.
	For example, a user wants to plan a trip to Beijing for relaxing, but he or she only has limited knowledge about Beijing. Thus it is difficult for him or her to decide which slots are needed to achieve this goal. Obviously, in this scene, the user needs additional consultant services from an agent to help figure out clear and specific goals. However, the aforementioned assumption hinders providing these services effectively.
		
\begin{figure*}[!ht]
		\centering
		\includegraphics[width=0.9\textwidth]{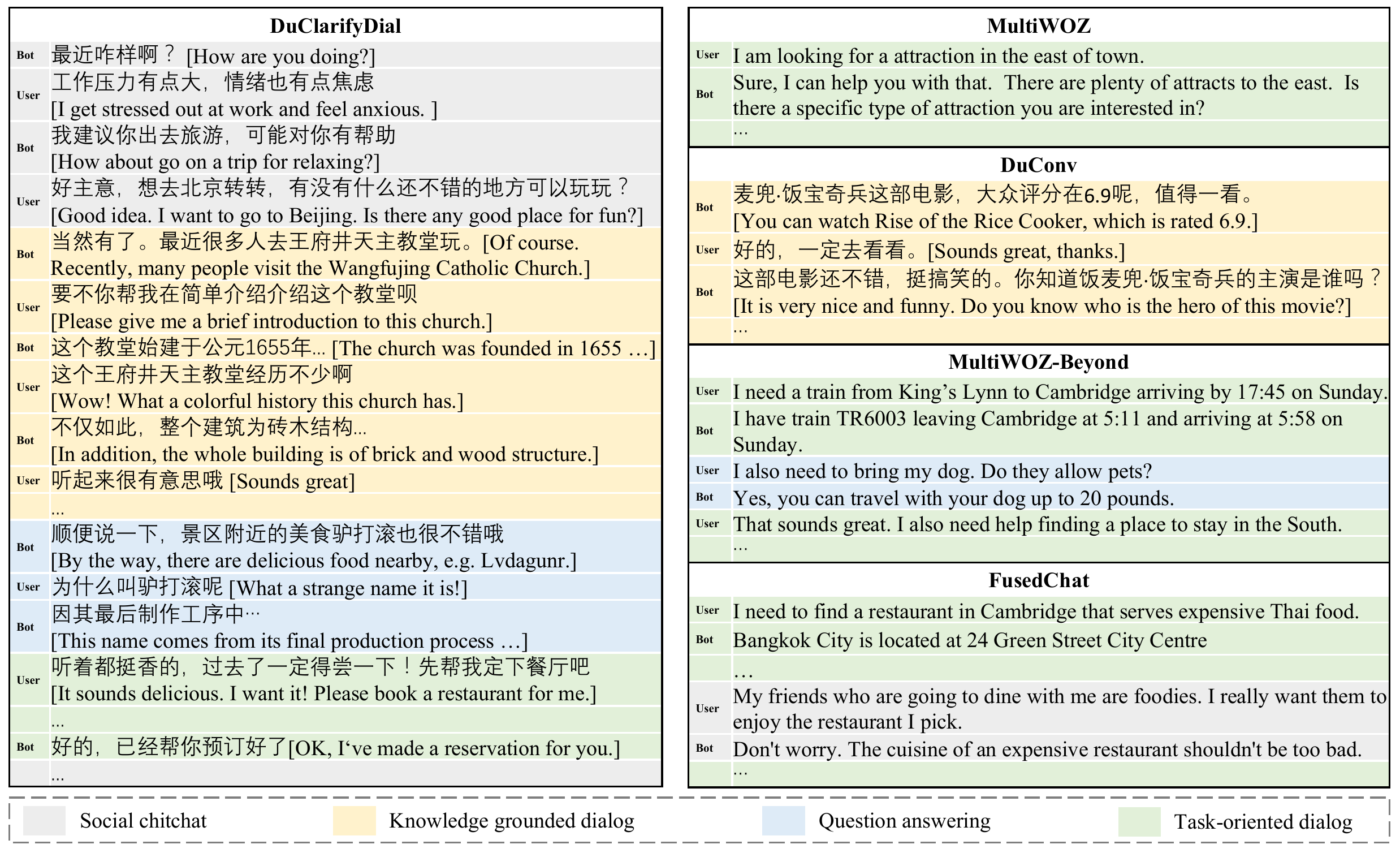}
		\caption{Dialog examples in DuClarifyDial and other dialog corpora. There are four dialog types in a single dialog session of DuClarifyDial while other dialog corpora contain one or two dialog types.}
		\label{fig_0}
	\end{figure*}

	In this paper, we make a step towards solving the challenge. In order to facilitate the study of how to help users \textbf{clarify}ing their goals, we construct a new \textbf{Dial}og corpus at Bai\textbf{du}, denoted as \textbf{DuClarifyDial}. \footnote{\url{https://github.com/PaddlePaddle/Research/tree/master/NLP/ACL2022-DuClarifyDial}}
	As shown in Figure~\ref{fig_0}, a user chats about ``feels anxious'' because of work pressure, and wants to relax himself or herself but have no clear idea about the trip. In the scenario, the agent conducts knowledge-grounded dialogs and question answering conversations to help the user learn more about goal-related knowledge, which helps figure out clear and specific goals. Finally, the user determines to visit ``Wangfujing Catholic Church'' and books a restaurant nearby. Specifically, in DuClarifyDial, besides basic social chitchat, an agent should help users figure out clear and specific goals by providing goal-related knowledge through coherent knowledge-grounded dialogs and question answering (QA) conversations. Then, upon request, it should also conduct task-oriented dialogs to help achieve user goals. 
	
	To this end, we first collect a human-to-human mixed-type dialog dataset. It contains 5k dialog sessions and 168k utterances for 4 dialog types and 5 domains. Specifically, each session contains at least two of following four dialog types, i.e., social chitchat, question answering, knowledge-grounded dialog, and task-oriented dialog.
	Furthermore, in order to seamlessly blend different types of dialogs, we make efforts in both dataset collection and task definition. For \textit{dataset collection}, we first collect human-to-human dialogs within the Wizard-of-Oz framework~\cite{kelley1984iterative}. Then, we design a unified dialog state schema and dialog act schema for all types of dialogs. Here, the unification can (1) ease the dialog annotation procedures, (2) simplify dialog model design, and (3) facilitate wiser dialog management by bringing a shared dialog semantic space for different types of dialogs. Finally, we annotate dialog states and dialog acts. 
	For \textit{task definition}, we first unify the dialog modelling into three sub-procedures, which includes dialog state tracking, dialog act planning and response generation. Then, we define one sub-task for each sub-procedure. Besides, in order to facilitate end-to-end modelling, we also define an end-to-end dialog generation sub-task. 

    To facilitate model comparison, we conduct bench-marking experiments on DuClarifyDial for the aforementioned four sub-tasks. 
     Furthermore, since DuClarifyDial is a mixed-type dialog corpus, it is straightforward to explore effective methods for utilizing existing single-type or mixed-types dialog corpora in task modelling. Specifically, we propose a novel Prompt-based continual learning mechanism to strengthen the model ability, by continually utilizing existing different types of dialog corpora. Here, we equip a pre-trained dialog model~\cite{Bao2020PLATOPD} with (1) different prompt texts as input and (2) type, task and domain representation in embedding layer for different dialog types. Furthermore, we train our model by two steps with continual learning mechanism: first Prompting on existing dialog corpora and then fine-tuning on DuClarifyDial.
	
	This work makes the following contributions: 
	\begin{itemize}
		\item We identify a new challenge that users have difficulties to figure out all the aspects of their goals in many real-world scenarios.
		\item We propose a large-scale Chinese mixed-type corpus, where each session weaves together multiple types of dialogs with natural cross-type transitions. Specifically, we design a unified dialog state (act) schema for all types of dialogs. Here, the unified organization first brings a shared semantic space for task-oriented and non-task-oriented dialogs. Then, it enables a unified dialog modelling procedures for all types of dialogs, which can facilitate more effective dialog management. 
		\item We build benchmarking baselines on DuClarifyDial and propose a  novel Prompt-based continual learning mechanism to utilize existing dialog corpora effectively.
	\end{itemize}

\section{Related Work}
\subsection{Multi-Domain Task-Oriented Dialog Datasets}
Task-oriented dialog systems have continued an active research area for decades and have been consistently supported by the development of new datasets. Recently, several large-scale multi-domain task-oriented dialog datasets have emerged~\cite{Budzianowski2018MultiWOZA,quan2020risawoz,Rastogi2020TowardsSM,Zhu2020CrossWOZAL,jin2021towards,Chen2021ActionBasedCD}. Specifically, MultiWOZ~\cite{Budzianowski2018MultiWOZA} is a fully-labelled collection of human-human written conversations spanning over multiple domains and topics, which contains a size of 10k dialogs. Schema~\cite{Rastogi2020TowardsSM} proposes a schema-guided paradigm for task-oriented dialog, which contains over 16k multi-domain conversations spanning 16 domains. CrossWOZ~\cite{Zhu2020CrossWOZAL} and RiSAWOZ~\cite{quan2020risawoz} are Chinese cross-domain task-oriented datasets, which contains 6K and 11k dialogs respectively. ABCD~\cite{Chen2021ActionBasedCD} includes over 10K dialogs that incorporate procedural, dual-constrained actions.

Although achieved promising progress, these datasets usually posit that users have figured out clear and specific goals before staring an interaction, which is not hold in many practical scenarios. In this paper, we focus on providing additional consultant services for users, to help figure out clear and specific user goals.

\subsection{Knowledge grounded Dialog Datasets}
Open-domain dialog systems have attracted lots of interests in recent years. To develop more human-like dialog models, several knowledge-grounded corpora have been proposed~\cite{wu2019proactive,Moon2019OpenDialKGEC,Liu2020TowardsCR,Zhou2020KdConvAC,Wang2021NaturalConvAC,Komeili2021InternetAugmentedDG,feng-etal-2020-doc2dial,yoshino2015conversational,tanaka2021arta}. The main purpose on these datasets is to generate more knowledgeable dialogs. In comparison, DuClarifyDial focuses on helping figure out clear and specific user goals. Moreover, DuClarifyDial is a mixed-type dialog dataset that contains four types of dialogs.

\subsection{Multi-tasking Dialogs}
Recently, there are multiple efforts on developing dialog systems that can multi-task on multiple types of dialogs~\cite{kim2020beyond,Smith2020CanYP,mosig2020star,madotto2020attention,saha2018towards,Sun2021AddingCT,Young2021fusing}. Specifically, Kim \textit{et al.}~\cite{kim2020beyond} propose to handle out-of-API requests, by accessing unstructured domain knowledge in task-oriented dialogs. Sun \textit{et al.}~\cite{Sun2021AddingCT} and Yong \textit{et al.}~\cite{Young2021fusing} propose to fuse task-oriented and open-domain dialogs in conversational agents, in order to generate more engaging and interactive dialogs.

The DuClarifyDial dataset differs from these datasets in that we focus on helping figure out clear and specific user goals, rather than targeting at the out-of-API problem ~\cite{kim2020beyond} or facilitating a more engaging and interactive dialog generation~\cite{Young2021fusing}. Furthermore, DuClarifyDial contains more types of dialogs than previous datasets. Moreover, in order to 
seamlessly blend different types of dialogs for efficient consulting, 
DuClarifyDial utilizes the same dialog state schema and dialog act schema for all types of dialogs, rather than utilizes different schema for different types of dialogs.

\section{The DuClarifyDial Dataset}
DuClarifyDial is designed to collect a high quality mixed-type dialog dataset for helping figure out clear and specific goals.
In DuClarifyDial, one person serves as the user and the other as the wizard (agent). In order to help figure out clear and specific goals, besides social chitchat, the agent provides user-goal-related information through knowledge grounded dialogs and QA conversations, and then help achieve the goals through task-oriented dialogs. 

Specifically, in order to effectively weave together multi types of dialogs for achieving this purpose, it is essential for different types of dialogs to share the same state space and action space. Thus, in Section \ref{annotation}, we utilize a unified dialog state schema and dialog act schema for the aforementioned four types of dialogs.

In the following, we will introduce the four steps of DuClarifyDial collection: (1) building knowledge base to provide goal-related information; (2) constructing dialog templates to assist dialog collection; (3) collecting conversation utterances by crowdsourcing; (4) annotating dialog states and dialog acts.

\begin{table}[t]
	\centering
	\resizebox{0.45\textwidth}{!}{
	\begin{tabular}{ p{0.9in}  p{1.9in} } 
		\toprule[1.0pt]
		Sub-Scena. & Description\\  
		\toprule[1.0pt]
		Sub-1: chitchat (Greeting)
		&The user says that his life was very monotonous.
		\\\hline
		Sub-2: chitchat(Help decision-making)
		&Bot suggests users travel. The user doesn't know where to go. Bot suggests the user go to Beijing. User consent.
		\\\hline
		Sub-3: Task-oriented dialog (Seek tourist attraction)
		&The user seeks for tourist attractions with high rating. Bot recommends the imperial palace, but the user has been there. Then, bot recommends users to fragrant hills. The user doesn't know fragrant hills.
		\\\hline
		Sub-4: Knowledge-grounded dialog (about fragrant hills)
		&The bot and user conduct an in-depth knowledge-grounded dialog about fragrant hills. Finally, the user wants to book tickets.
		\\\hline
		Sub-5: Task-oriented dialog (Book tickets)
		&Bot helps the user book tickets.
		\\
		\bottomrule[1.0pt]
	\end{tabular}
	}
	\caption{An example dialog template.}
	\label{table:template}
\end{table}

\subsection{Knowledge Base Construction}
In order to create a knowledge base that includes five domains: hotel, attraction, restaurant, food, and movie, we collect publicly available information from the WEB. Specifically, for the hotel domain, we collect 1,133 entities and their related knowledge from two famous online accommodation reservation websites, Qunar and Ctrip. \footnote{https://www.qunar.com/}\footnote{https://www.ctrip.com/} For the attraction domain, we collect 435 entities and their related knowledge from the famous travelling website, Mafengwo. \footnote{http://www.mafengwo.cn/} For the restaurant domain, we collect 122 entities and their related knowledge from the famous shopping platform, Meituan. \footnote{https://www.meituan.com/} For the  food domain, we collect 1,971 entities and their related knowledge from the famous online encyclopedia, Baidu Baike. \footnote{https://baike.baidu.com/} Finally, for the movie domain, we collect 224 entities and their related knowledge from two famous social networking websites, Mtime and Douban. \footnote{http://www.mtime.com/}\footnote{https://www.douban.com/}

\subsection{Dialog Template Construction}
Based on the collected knowledge base, we generate dialog templates to guide crowdsourcing workers, which is in line with previous work\cite{Budzianowski2018MultiWOZA,Liu2020TowardsCR}. Here, each template consists of a sequence of dialog sub-scenarios, and each sub-scenario is defined by a dialog type, a dialog topic and a detailed description text. Table \ref{table:template} shows an example dialog template. Specifically, in order to better imitate the real scenarios, dialog templates should introduce different interaction behaviours. For example, a user may ask for reserving a ticket during conducting an in-depth knowledge-grounded dialogs around a certain entity, e.g., an attraction. Furthermore, a user may interrupt a task-oriented dialog by chatting about some instant content in mind, and then continue the task-oriented dialog. 

In order to construct dialog templates, we first utilize heuristic rules to automatically enumerate candidate sub-scenarios sequences that have natural topic transitions. Then, we utilize pre-defined templates to generate detailed descriptions for these sub-scenarios. Finally, to further ensure natural topic transitions, we manually filter out a few incoherent dialog templates, such as descriptions that contain inconsistent facts. 

\subsection{Dialog Collection}
	In order to collect high quality dialogs, we set a strict annotation procedure to guide workers to annotate dialogs based on the given templates. Specifically, the collection procedure includes three stages: (1) reliable crowdsourcing workers recruitment, (2) dialog generation, and (3) quality verification.
	
	\begin{figure}[!ht]
		\centering
		\includegraphics[width=0.48 \textwidth]{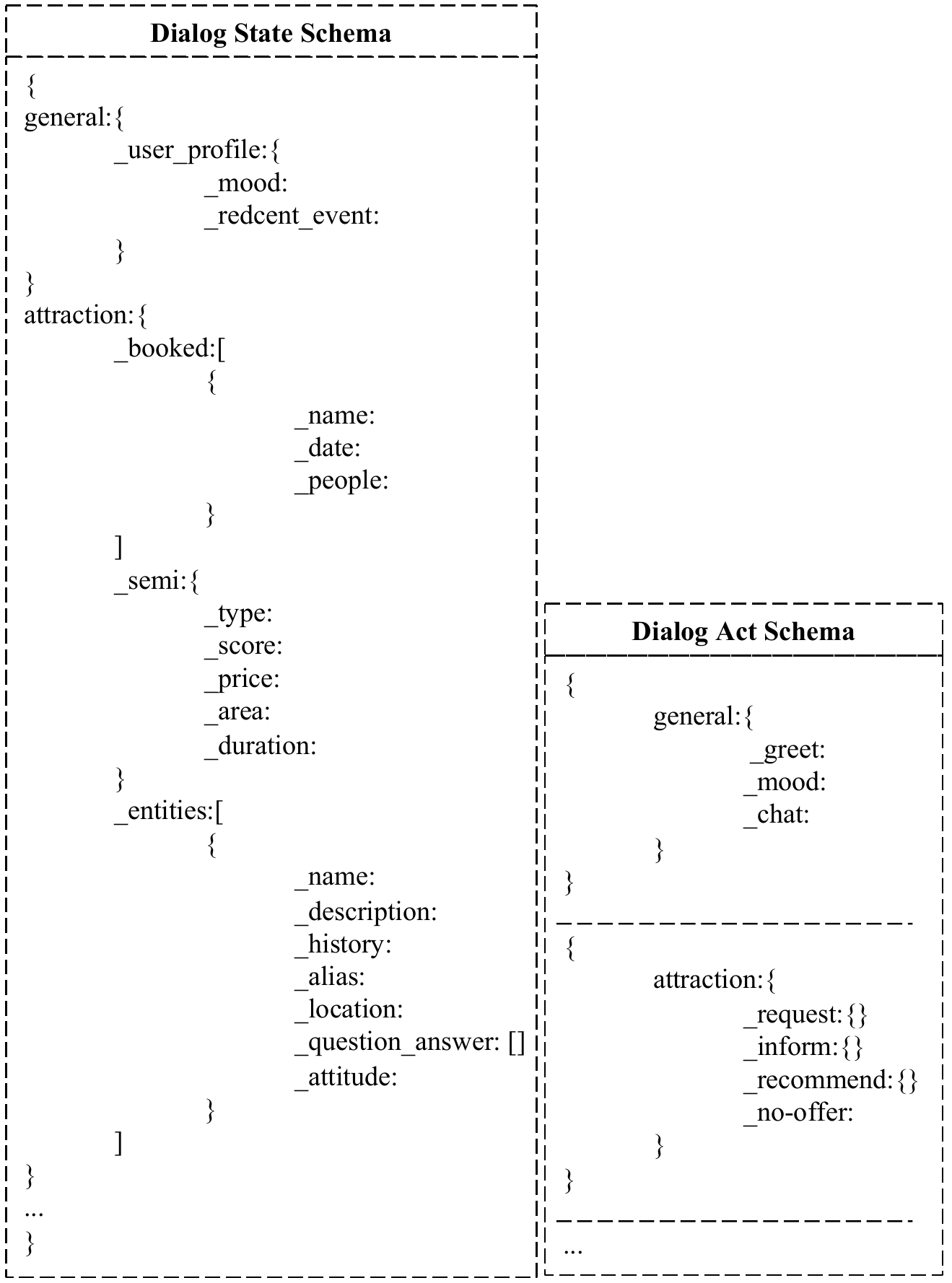}
		\caption{The unified dialog state schema and dialog act schema.}
		\label{fig_4}
	\end{figure}
 
	\textbf{In the worker recruitment stage}, in order to select reliable workers, we recruit 100 candidates in a famous crowdsourcing platform. \footnote{https://test.baidu.com/} Then, we ask each candidate to label 10 dialog sessions based on given templates. Lastly, we employ the top-40 candidates with the highest labelling quality to serve as crowdsourcing workers.

	\textbf{In the dialog generation stage}, we develop a labelling interface for crowdsourcing workers to converse synchronously. 
	Then, we randomly pair up two crowdsourcing workers and set each of them a role of the user or the wizard (bot). Lastly, the two crowdsourcing workers generate dialogs with the help of the aforementioned knowledge base and dialog templates. 
	
	\textit{User Side} For a given dialog template, in order to prevent information overload, we only provide a sub-scenario to the user at a time. During dialog collection, a user first reads though the detailed description to understand the provided sub-scenario. Then, based on the given sub-scenario, the user communicates with the wizard turn by turn. Finally, the user may require for another sub-scenario if he or she believes the current sub-scenario has been accomplished. Specifically, in order to diversify the corpus, we encourage the users to follow their own speaking style in communication.

	\textit{Wizard Side} A wizard is required to serve as a consultant, who is responsible for helping users figure out clear and specific goals. At each sub-scenario, the wizard can get access to the associated knowledge in the interface, which is extracted from the knowledge base automatically. When receiving an utterance from the user side, the wizard needs to respond appropriately.

	\textbf{In the quality verification stage}, we manually check the collected dialogs. Specifically, if a dialog is considered as unqualified, we will ask the two crowdsourcing workers to revise the dialog until it is qualified.

\subsection{Dialog Annotation}
\label{annotation}
After collecting the conversation data, we recruit crowdsourcing workers to annotate dialog states and dialog acts. Specifically, in order to seamlessly blend multi types of dialogs for helping users figure out clear and specific goals, we first design a unified dialog states schema and dialog act schema for all types of dialogs, and then annotate the dialogs based on the schema.

\textbf{The unified dialog state} consists of a list of domain-states, as shown in Figure \ref{fig_4}. Specifically, we add a ``general'' domain to store user-profile related states, e.g., user mood. The ``general'' domain is important, since user-profile may have a significant impact on his or her goal. For other domains, we split domain-states into three parts: (1)``\_booked'' for storing booked orders in this domain. Each booked order contains all the necessary information for finishing the order; (2) ``\_semi'' for storing the important but not necessary information for an order; (3) ``\_entities'' for storing all the mentioned entities and the mentioned specific pieces of information about these entities. Specifically, we store an ``\_attitude'' slot in each mentioned entity to capture user interest directly. The values of the ``\_attitude'' slot contain two types: positive and negative. Here, the ``\_booked'' part is mainly corresponding to the task-oriented dialog, the ``\_entities'' part is mainly corresponding to the knowledge grounded dialog and question answering dialog, and the ``\_semi'' part corresponding to all the aforementioned three dialog types. 

\begin{figure*}[!ht]
	\centering
	\includegraphics[width=\textwidth]{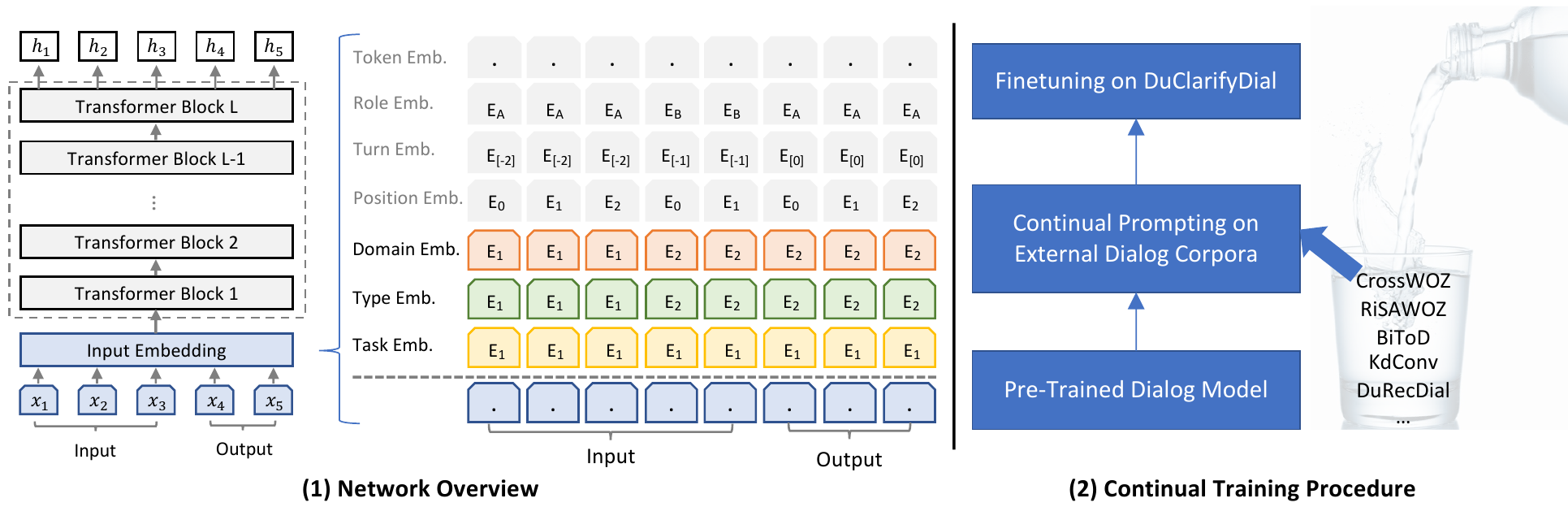}
	\caption{Overview of PLATO-MT}
	\label{fig_3}
\end{figure*}

\textbf{The unified dialog act schema} consists of domains, intents, slots and values. Specifically, we add a ``general'' domain to store intents that are not directly related to user goals. For other domains, they usually contain four intents: ``\_request'', ``\_inform'', ``\_recommend'' and ``\_no-offer''. Specifically, the classical knowledge selection in knowledge-ground dialog is treated as an ``\_inform'' action in this unified act schema.

\begin{table}[!ht]
	\centering
	\resizebox{0.4\textwidth}{!}{
		\begin{tabular}{p{2cm}lll}
			\toprule
			& \textbf{Train}  & \textbf{Dev}  & \textbf{Test}\\
			\midrule
			\# Dialogs & 3,500 & 500 & 1,052  \\
			\#Utt. & 117,301 & 16,543 &  34,999 \\
			Avg. utt. per dialog & 33 & 33 & 33 \\
			\#Tokens & 1,181,669 & 168,030 & 352,510 \\
			Avg. tokens per utt. & 10 & 10 & 10  \\
			\# Chitchat & 3,500 & 500 & 1,052 \\
			\# Know.  & 3,500 & 500 & 1,052 \\
			\# Task  & 3,500 & 500 & 1,052 \\
			\# QA  & 214 & 24 & 1,052 \\
			\bottomrule
		\end{tabular}
	}
	\caption{Dataset statistics.}
	\label{fig:dataset_statistics} 
\end{table}

Based on the unified schema, we recruit 10 crowdsourcing workers to annotate these dialog states and dialog acts. Specifically, before formal annotation, each worker must pass a labelling test. Here, we first annotate 10 dialogs manually. Then, we ask workers to annotate these dialogs. Lastly, a worker passes the test if his annotations are the same as our annotations.

\subsection{Overall Dataset}

The overall collected data consists of 5,052 dialog sessions in total, with 3,000 sessions in the training set, and validation and test sets of 500 and 1,052 sessions, respectively. Overall statistics can be found in Table 2. 
		
We conduct human evaluations for data quality. Specifically, if a dialog follows the instruction in task templates and the utterances are fluent and grammatical, it will be rated ``1'', otherwise ``0''. Then we ask three workers to judge the quality of 200 randomly sampled dialogs. Finally we obtain an average score of 0.83 on this evaluation set.

\section{The Mixed-Type Dialog Model with Prompt-based Mechanism}
Recently, large scale pre-trained dialog models have achieved impressive performance, both in task-oriented dialog~\cite{Heck2020TripPyAT,Yang2021UBARTF} and open-domain chitchat~\cite{adiwardana2020towards,roller2021recipes,Bao2020PLATOPD}. Meanwhile, the methodologies for different types of dialogs have gradually shifted to generative and end-to-end modelling. Following these trends, we propose a pre-trained mixed-type dialog model based on ~\cite{Bao2020PLATOPD}, denoted as PLATO-MT. Furthermore, we equip our model with a novel Prompt-based continual learning mechanism to strengthen the model ability by continually utilizing external existed different types of dialog corpora.

\subsection{The Prompt-based Continual Learning Mechanism}
Figure~\ref{fig_3} shows an overview of the proposed PLATO-MT model. As shown in Figure~\ref{fig_3} (1), the model is a multi-layer transformer-based neural network. Furthermore, the inputs and outputs of all dialog sub-tasks are formalized as simple text sequences. 

In order to effectively blend the abilities of mixed-type dialog in one model, we follow the ``Prompt + LM Fine-tuning'' strategy~\cite{liu2021pre}. Specifically, we design different Prompt texts as input for different dialog types. For example, for knowledge-based dialogs, the Prompt text of input is ``[Knowledge] context''. Here ``[Knowledge]'' refers to knowledge sentences used for context. Similarly, the Prompt text of QA is ``[Question|Answer] context'' and the Prompt text of task-oriented dialog is ``[Domain|Slot|Value] context''. 
Furthermore, we add type, task and domain embedding representation in embedding layers to further differentiate the characters of different dialog types.

Meanwhile, we train the PLATO-MT model with continuous learning mechanism, as shown in Figure~\ref{fig_3} (2). In particular, we first carry on prompting on existing dialog corpora, such as CrossWOZ~\cite{Zhu2020CrossWOZAL}, RiSAWoz~\cite{quan2020risawoz}, BiToD~\cite{lin2021bitod},  Kdconv~\cite{Zhou2020KdConvAC} and DurecDial~\cite{Liu2020TowardsCR}. Thus we strengthen our model ability by continually utilizing external existed different types of dialog corpora.
Then we finetune the prompted model on our proposed dialog corpus DuClarifyDial.

\begin{table*}[!ht] 
	\centering
		\begin{tabular}{l c  c c  c c  c c c}
			\hline
			\multicolumn{1}{l|}{\multirow{2}{*}{\bf Methods}} &  \multicolumn{4}{c|}{\bf Sub-Task1: DST} & \multicolumn{2}{c}{\bf Sub-Task2: DAP} \\ \cline{2-7} \multicolumn{1}{l|}{} 
			& \multicolumn{1}{|l}{Type Acc.} &  Domain Acc. & Slot Acc. & \multicolumn{1}{c|}{Joint Acc.}  & Act Acc. &  \multicolumn{1}{c}{BLEU-1/2} \\
			\hline
			\multicolumn{1}{l|}{UBAR}  & 0.96 & 0.95  & 0.77  & \multicolumn{1}{c|}{0.39} & 0.85 &\multicolumn{1}{c}{0.84/0.83} \\
			\multicolumn{1}{l|}{MinTL}  & \textbf{0.99}  & 0.94  & 0.86  & \multicolumn{1}{c|}{0.48} & 0.87 &\multicolumn{1}{c}{0.83/0.82} \\
			\multicolumn{1}{l|}{PLATO}  & \textbf{0.99}  & \textbf{0.97} & 0.85  & \multicolumn{1}{c|}{0.48} & 0.91 &\multicolumn{1}{c}{0.89/0.88} \\
			\multicolumn{1}{l|}{PLATO-MT} & \textbf{0.99}  & \textbf{0.97}  & \textbf{0.88}  & \multicolumn{1}{c|}{\textbf{0.51}} & \textbf{0.93} &\multicolumn{1}{c}{\textbf{0.90}/\textbf{0.90}} \\
			\multicolumn{1}{l|}{-w/o Prompt} & \textbf{0.99}  & \textbf{0.97}  & 0.87  & \multicolumn{1}{c|}{0.49} & 0.92  &\multicolumn{1}{c}{\textbf{0.90}/0.89} \\
			\hline
		\end{tabular}
	\caption{DST and DAP Results on DuClarifyDial.}
	\label{tab3}
\end{table*}

\begin{table*}[ht] 
	\centering
	\resizebox{\textwidth}{!}{
		\begin{tabular}{l c  c c  c c  c c c c}
			\hline
			\multicolumn{1}{l|}{\multirow{2}{*}{\bf Methods}} &  \multicolumn{4}{c|}{\bf Automatic Metrics} &  \multicolumn{4}{c}{\bf Manual Metrics} \\ \cline{2-9} \multicolumn{1}{l|}{} 
			& BLEU-1/2  & METEOR & CIDER &  \multicolumn{1}{c|}{Dist-1/2} & Appr. & Info. & Hallu. &  Suc.\\
			\hline
			\multicolumn{1}{l|}{UBAR} & 0.39/0.32 & 0.21 & 2.28 &\multicolumn{1}{c|}{0.006/0.040} & 0.88 & 0.91  & 0.45 & 0.43 \\
			\multicolumn{1}{l|}{MinTL} & 0.37/0.32 & 0.21 & 2.50 &\multicolumn{1}{c|}{0.007/0.079} & 0.91 & 0.93  & 0.89 & 0.91  \\
			\multicolumn{1}{l|}{PLATO} & 0.46/0.39  & 0.25 & 2.57  &\multicolumn{1}{c|}{0.007/0.072} & 0.97 & 0.95  & 0.90 & 0.90 \\
			\multicolumn{1}{l|}{PLATO-MT} & \textbf{0.50}/\textbf{0.43} & \textbf{0.27} & \textbf{3.00}  &\multicolumn{1}{c|}{\textbf{0.008}/\textbf{0.083}} & \textbf{0.99}  & \textbf{0.97} & \textbf{0.93} & \textbf{0.94}   \\
			\multicolumn{1}{l|}{-w/o Prompt} & 0.46/0.40 & 0.26 & 2.84  &\multicolumn{1}{c|}{\textbf{0.008}/0.079} & 0.93  & 0.96 & 0.90 & 0.90  \\
			\hline
		\end{tabular}
	}
	\caption{RG Results on DuClarifyDial.}
	\label{tab4}
\end{table*}

\begin{table*}[ht] 
	\centering
	\resizebox{\textwidth}{!}{
		\begin{tabular}{l c  c c  c c  c c c c}
			\hline
				\multicolumn{1}{l|}{\multirow{2}{*}{\bf Methods}} &  \multicolumn{4}{c|}{\bf Automatic Metrics} &  \multicolumn{4}{c}{\bf Manual Metrics} \\ \cline{2-9} \multicolumn{1}{l|}{} 
				& BLEU-1/2  & METEOR & CIDER &  \multicolumn{1}{c|}{Dist-1/2} & Appr. & Info. & Hallu. &  Suc.\\
				\hline
				\multicolumn{1}{l|}{UBAR} & 0.28/0.22 & 0.16 & 1.70 &\multicolumn{1}{c|}{0.005/0.031} & 0.74 & 0.87  & 0.32 & 0.34  \\
				\multicolumn{1}{l|}{MinTL} & 0.32/0.25 & 0.17 & 1.80 &\multicolumn{1}{c|}{0.006/0.046} & 0.86 & 0.88  & 0.35 & 0.34  \\
				\multicolumn{1}{l|}{PLATO} & 0.32/0.25  & 0.16 & 1.28  &\multicolumn{1}{c|}{0.005/0.034} & 0.78 & 0.88  & 0.36 & 0.36 \\
				\multicolumn{1}{l|}{PLATO-MT} & \textbf{0.45}/\textbf{0.37} & \textbf{0.23} & \textbf{2.17}  &\multicolumn{1}{c|}{\textbf{0.007}/\textbf{0.072}} & \textbf{0.96}  & \textbf{0.90} & \textbf{0.67} & \textbf{0.69}  \\
				\multicolumn{1}{l|}{-w/o Prompt} & 0.41/0.33 & 0.21 & 1.89  &\multicolumn{1}{c|}{\textbf{0.007}/0.062} & 0.87  & 0.89 & 0.55 & 0.52  \\
				\hline
		\end{tabular}
	}
	\caption{E2E-DG Results on DuClarifyDial.}
	\label{tab5}
\end{table*}

\section{DuClarifyDial as a New Benchmark}
We break down the mixed-type dialog modelling task into three sub-tasks: dialog state tracking, dialog act planning, and dialog-act-to-text generation. Besides, in order to facilitate end-to-end dialog modelling, we define an end-to-end dialog-context-to-text generation sub-task. For each of the four sub-tasks, we report benchmark results on the following dialog models, which have achieved promising performance in the popular MultiWOZ dataset~\cite{Budzianowski2018MultiWOZA}. Specifically, we use the original codes released by the authors. 

\noindent \textbf{UBAR}~\cite{Yang2021UBARTF} UBAR is a fully end-to-end task-oriented dialog model that takes a pre-trained model as backbone. Here, since DuClarifyDial is a Chinese dataset, we utilize a Chinese large-scale pre-trained model, ERNIE~\cite{xiao2020ernie-gen}, to initialize UBAR.
	
\noindent \textbf{MinTL}~\cite{Lin2020MinTLMT} MinTL is a strong model that utilizes effective transfer learning to plug-and-play pre-trained models. Here, instead of utilizing BART~\cite{lewis2020bart} as in the original paper, we utilize the multi-lingual version,  mBART~\cite{liu2020multilingual}, for initialization.

\noindent \textbf{PLATO}~\cite{Bao2020PLATOPD} PLATO is the state-of-the-art Chinese pre-trained dialog model. We use the released parameters. \footnote{https://github.com/PaddlePaddle/Knover/tree/luge-dialog/luge-dialog}

\noindent \textbf{PLATO-MT} It is the proposed unified mixed-type dialog model with Prompt-based Continual Learning mechanism. Here, the Prompt-related parameters are random initialized.

\noindent \textbf{PLATO-MT w/o Prompt} It is the PLATO-MT model without Prompting. We first fine-tune it on the same set of existing dialog corpus as in PLATO-MT, and then fine-tune it on DuClarifyDial.

\subsection{Dialog State Tracking}
For building a successful dialog system, a robust dialog state tracking (\textbf{DST}) is considered as the first step. It takes previous dialog utterances and the recent dialog state as input, and then outputs the current dialog state. 

To evaluate the performance on dialog state tracking, we utilize both slot-level metric and dialog-level metrics. For slot-level metric, we measure the slot accuracy (\textbf{Slot Acc.}). Specifically, the slot accuracy is measured by individually comparing each (domain, slot, value) triplet to its ground truth label. For dialog-level metric, besides dialog type accuracy (\textbf{Type Acc.}) and dialog domain accuracy (\textbf{Domain Acc.}), we also measure the joint goal accuracy (\textbf{Joint Acc.})~\cite{wu2019transferable}. It compares the predicted dialog states to the ground truth at each turn, and the output is considered correct if and only if all the predicted values exactly match the ground truth. 

Table~\ref{tab3} shows the evaluation results. We can see all the models achieve promising results in terms of ``Type Acc.'' and ``Domain Acc.''. It indicates the effectiveness of utilizing large-scale pre-trained models as backbone. Furthermore, we notice that PLATO-MT outperforms all the baselines, especially in terms of ``Slot Acc.'' and ``Joint Acc.''. It demonstrates that PLATO-MT can track dialog states effectively.

\subsection{Dialog Act Planning}
\label{DAP}
The dialog act planning (\textbf{DAP}) sub-task takes dialog context, current dialog state and retrieved coarse knowledge as input, and then outputs system act.
Specifically, for each dialog session, we first extract all the entities in it, and then retrieve all the related knowledge about these entities to serve as the retrieved coarse knowledge. 

To evaluate the performance on dialog act planning, we measure the dialog act accuracy (\textbf{Act Acc.}) and the BLEU-1/2~\cite{papineni2002bleu} score.

Table~\ref{tab3} shows the evaluation results. We notice that PLATO-MT outperforms all the baselines, especially in terms of ``Act Acc.''. It demonstrates that PLATO-MT can plan appropriate dialog acts effectively.

\subsection{Dialog-Act-to-Text Generation}
\label{context-to-generation}
The dialog act to text generation (\textbf{RG}) sub-task aims to transform a structured dialog act into a response. It takes dialog context and delexicalized dialog act as input, and then outputs a response. 

To evaluate performance on generation, we utilize both automatic metrics and manual metrics. For automatic evaluation, we use several classical metrics, including BLEU~\cite{papineni2002bleu}, METEOR~\cite{banerjee2005meteor}, CIDER~\cite{vedantam2015cider} and Distinct (\textbf{Dist.})~\cite{li2016diversity}.

For manual evaluation, we conduct evaluation on randomly sampled 50 sessions at the level of both turns and dialogs. For turn-level human evaluation, the generated responses are evaluated by three annotators in terms of appropriateness (\textbf{Appr.}) and informativeness (\textbf{Info.}). For dialog-level human evaluation, we measure hallucination (\textbf{Hallu.}) that measures information accuracy in generated responses, and dialog success (\textbf{Suc.}) that measures whether an agent helps users figure out clear goals. Specifically, if a user has not completed any order during a session, the success score is 0; Otherwise, the success score equals to the information accuracy in a session. 

Table~\ref{tab4} shows the evaluation results. We find PLATO-MT significantly outperforms all the baselines in terms of all the metrics except ``Dist-1/2'' (sign test, p-value < 0.01). It indicates that PLATO-MT can generate dialogs with higher qualities.

\subsection{End-to-End Dialog Generation}
This end-to-end dialog generation sub-task (\textbf{E2E-DG}) takes dialog context as input, and then outputs an utterance for responding. Specifically, in the end-to-end settings, since the dialog domain and type information are not available at each turn, we do not use them as input information. Here, we consider the same set of evaluation settings as in Section\ref{context-to-generation}.

Table~\ref{tab5} shows the evaluation results. We find PLATO-MT significantly outperforms all the baselines in terms of all the metrics except ``Dist-1/2'' (sign test, p-value < 0.01). Specifically, in terms of ``Hallu.'' and ``Suc.'' in manual evaluation, PLATO-MT outperforms other models by a large margin. It indicates that PLATO-MT is much more competent in helping users learn about correct goal-related knowledge, which is essential for helping users figure clear and specific goals. 

\subsection{Ablation Study}
In order to evaluate the contribution of the proposed Prompt-based continual learning mechanism, we remove the mechanism from PLATO-MT, denoted as ``PLATO-MT-w/o Prompt''. Here, we first fine-tune PLATO on the same set of existing dialog corpus as in PLATO-MT, and then fine-tune it on DuClarifyDial. For evaluation, we consider the same set of settings as in Section\ref{context-to-generation}.

As shown in Table~\ref{tab3}, Table~\ref{tab4} and Table~\ref{tab5}, its performance drops in terms of most metrics in all the four sub-tasks.
Specifically, in manual evaluation in Table~\ref{tab5}, we notice a sharp performance degradation in terms of  ``Hallu.'' and ``Suc.''. It demonstrates the Prompt-based mechanism is essential for effectively utilizing existing dialog corpora, which enables PLATO-MT can continually strengthen its ability on any specific dialog type. 

Furthermore, we find that, in terms of most metrics, the mechanism gains more in the end-to-end conversation generation sub-task than in the other three sub-tasks. This is because there are no available annotated information in the end-to-end conversation generation sub-task, which makes it a more difficult task. Thus, the effect of Prompt-based continual mechanism appears relatively more significant. 
		
\section{Conclusion}
In this paper, we first identify the challenge that users may struggle to figure out clear and specific goals in many real scenarios. Then, we make a step forward by collecting a new human-to-human mixed-type dialog corpus, which contains 5k dialog sessions and 168k utterances for 4 dialog types and 5 domains. Furthermore, we setup benchmarks based on the corpus. Moreover, we propose a mixed-type dialog generation model with a novel Prompt-based continual learning mechanism. Finally, experimental results demonstrate the effectiveness of the mechanism.

\section* {Ethical Considerations}
We make sure that \emph{DuClarifyDial} has been collected in a manner that is consistent with the terms of use of any sources and the intellectual property and privacy rights of the original authors of the texts. 
And crowd workers were treated fairly. This includes, but is not limited to, compensating them fairly and ensuring that they were able to give informed consent, which includes, but is not limited to, ensuring that they were voluntary participants who were aware of any risks of harm associated with their participation.
Please see Section 3 for more details characteristics and collection process of \emph{DuClarifyDial}.

\bibliography{acl2022_new_final.bib}
\bibliographystyle{acl_natbib.bst}
	
\end{document}